\documentclass[letterpaper]{article}
\usepackage{aaai}
\usepackage{times}
\usepackage{helvet}
\usepackage{courier}
\usepackage{subfigure, float, graphicx, epstopdf, balance}
\usepackage{color, soul}
\usepackage[table]{xcolor}
\usepackage{epsfig}
\usepackage{booktabs}

\usepackage{mathtools, url}
\usepackage{latexsym}
\usepackage{amsfonts} 
\usepackage{adjustbox}
\usepackage{lipsum}

\definecolor{lightgray}{gray}{0.85}

\begin{document}

\title{Non-linear Label Ranking for Large-scale \\ Prediction of Long-Term User Interests}

\author{Nemanja Djuric$^\dag$, Mihajlo Grbovic$^\dag$, Vladan Radosavljevic$^\dag$, Narayan Bhamidipati$^\dag$, Slobodan Vucetic$^\ddag$\\
$^\dag$Yahoo! Labs, Sunnyvale, CA, USA, \{nemanja, mihajlo, vladan, narayanb\}@yahoo-inc.com\\
$^\ddag$Temple University, Philadelphia, PA, USA, vucetic@temple.edu\\
}



\maketitle
\begin{abstract}
We consider the problem of personalization of online services from the viewpoint of ad targeting, where we seek to find the best ad categories to be shown to each user, resulting in improved user experience and increased advertisers' revenue. We propose to address this problem as a task of ranking the ad categories depending on a user's preference, and introduce a novel label ranking approach capable of efficiently learning non-linear, highly accurate models in large-scale settings. Experiments on a real-world advertising data set with more than $3.2$ million users show that the proposed algorithm outperforms the existing solutions in terms of both rank loss and top-$K$ retrieval performance, strongly suggesting the benefit of using the proposed model on large-scale ranking problems.
\end{abstract}



\section{Introduction}
Personalization of online content has become an important topic in the recent years. It has been defined as "the ability to proactively tailor products and product purchasing experiences to tastes of individual consumers based upon their personal and preference information" \cite{chellappa2005personalization}, which may lead to improved user experience and directly translate into financial gains for online businesses \cite{riecken2000personalized}. In addition, personalization fosters stronger bond between users and companies, and can help in increasing user loyalty and retention \cite{alba1997interactive}. For these reasons it has been recognized as an important strategic goal of major internet companies \cite{manber2000yahoo,das2007google}, and is a focus of significant research efforts. Personalized content has already become an integral part of many popular online services, a trend likely to continue in the future \cite{tuzhilin2009personalization}.

We consider content personalization from the viewpoint of targeted advertising \cite{essex2009matchmaker}, an increasingly important aspect of online businesses. Here, for each individual user the task is to find the best matching ads to be displayed, which improves user's online experience (as only relevant and interesting ads are shown to the user) and can lead to increased revenue for the advertisers (as users are more likely to click on the ad and make a purchase). Due to its large impact and many open research questions, targeted advertising has garnered significant interest from the machine learning community, as witnessed by a large number of recent workshops and publications \cite{broder2008computational,pandey2011learning,majumder2013know}.

One of the most popular approaches in present-day targeting, particularly in brand awareness campaigns, is to assign categories to the display ads, such as "sports" or "finance", and then separately learn to predict user interest in each of these categories using historical records \cite{ahmed2011scalable,pandey2011learning,tyler2011retrieval}. Typically, a taxonomy is used to decide on the categories, and depending on how detailed it is hundreds of separate category qualification tasks may need to be solved. Thus, for each ad category, a separate predictive model is trained, able to estimate the probability of an ad click for the entire user population. Then, for each category, $N$ users with the highest click probability are selected for ad exposure. Known issues with the approach include overexposure, where a single user may be among the top $N$ users for many categories, and starvation, where some users do not qualify for any of the categories.


An alternative avenue, known in the industry as a user-interest model, is to sort for each user outputs of the predictive models, and qualify users based on their top $K$ categories. The approach guarantees that a user is qualified into several categories, eliminating overexposure and starvation issues. However, this method may still be suboptimal, as the predictive models are trained in isolation and do not consider relationships between different categories. In this paper we explore methods capable of capturing more complex class dependencies, and consider the user-interest model from a label ranking standpoint \cite{vembu2011label}. However, the sheer scale of ad targeting problems, with data sets comprising millions of users and features and hundreds of categories, renders many existing label ranking approaches intractable, presenting new challenges to the researchers. 

To address this issue, we propose a novel label ranking algorithm suitable for large-scale settings. The method lends ideas from the state-of-the-art AMM classifiers \cite{wang:djuric}, efficiently learning accurate, non-linear models on limited resources. Empirical evaluation was performed in a real-world ad targeting setting, using, to the best of our knowledge, the largest dataset considered thus far in the label ranking literature. The results show that the algorithm significantly outperformed the existing methods, indicating the benefits of the proposed approach to label ranking tasks.

\section{Background}
In this section we present works and ideas that led to the proposed algorithm. We first discuss label ranking setting, and then describe Adaptive Multi-hyperplane Machine (AMM), a non-linear, multi-class model used to develop a novel large-scale label ranking approach introduced in this paper.

\subsection{Label ranking}
Unlike standard machine learning problems such as multi-class or multi-label classification, label ranking is a relatively novel topic which involves a complex task of label preference learning. More specifically, rather than predicting one or more class labels for a newly observed example, we seek to find a strict ranking of classes by their importance or relevance to the given example. For instance, let us consider targeted advertising domain, and assume that the examples are internet users and class labels are user preferences from the set $\mathcal{Y} = \{$"sports", "travel", "finance"$\}$. Then, instead of simply inferring that the user is a "sports" person, which would result in user being shown only sports-related ads, it is more informative to know that the user prefers sports over finance over travel, resulting in more diverse and more effective ad targeting.
Note that the label ranking problem differs from the learning-to-rank setup \cite{LearningToRank}, where the task is to rank the examples and not labels, and can also be seen as a generalization of classification and multi-label problems \cite{Dekel}.

More formally, in the label ranking scenario the input is defined by a feature vector ${\bf x} \in \mathcal{X} \subset \mathcal{R}^d$, and the output is defined by a ranking $\pi \in \Pi$ of class labels. Here, the labels originate from a predefined set $\mathcal{Y} = \{1, 2, \ldots, L\}$ (e.g., $\pi=[3, 1, 4, 2]$ for $L = 4$), and $\Pi$ is a set of all label permutations. Let us denote by $\pi_i$ a class label at the $i^{\text{\scriptsize th}}$ position in the label ranking $\pi$, and by $\pi_i^{-1}$ a position (or rank) of label $i$ in the ranking $\pi$. For instance, in the above example we would have $\pi_1 = 3$ and $\pi_1^{-1} = 2$. Then, for any $i$ and $j$, where $0 \leq i < j \leq L$, we say that label $\pi_i$ is {\it preferred} over label $\pi_j$, or equivalently $\pi_i \succ \pi_j$. Moreover, in the case of an incomplete order $\pi$, we say that any label $i \in \pi$ is preferred over the missing ones.
Further, let us assume that we are given a sample from the underlying distribution $\mathcal{D} = \{({\it \ensuremath{{\bf x}_{t}}}, {\it \ensuremath{\pi_{t}}}), {\it t} = 1, ..., {\it T}\}$, where {\it \ensuremath{\pi_{t}}} is a vector containing either a total or a partial order of class labels $\mathcal{Y}$. The learning goal is to find a model $f$ that maps input examples ${\bf x}$ into a total ordering of labels, $f: \mathcal{X} \rightarrow \Pi$. 

In the recent years the problem has seen increased attention by the machine learning community (e.g., see recent workshops and tutorials at ICML, NIPS, and other venues), and many effective algorithms have been proposed in the literature \cite{Har-Peled,Dekel,Kamishima,Cheng1,grbovic2013}; for an excellent review see \cite{vembu2011label}. In \cite{Cheng1,Cheng2} authors propose instance-based methods for label ranking, where training examples are first clustered according to their feature vectors, and then centroid and mean ranking are found for each cluster and used for inference. This idea was extended in \cite{grbovic2013supervised,grbovic2013}, where authors use feature vectors to supervise clustering, resulting in improved performance. Apart from the prototype-based methods, often considered approaches include learning a scoring function $g_i$ for each class, $i = 1, \ldots, L$, and sorting their output in order to infer label ranking  \cite{elisseeff2001kernel,Dekel,Har-Peled}, or training a number of binary classification models to predict pairwise label preferences and aggregating their output into a total order \cite{hullermeier2008label,hullermeier2010combining}.

\subsection{Adaptive Multi-hyperplane Machine}
\label{sect:amm}
The AMM algorithm is a budgeted, multi-class method suitable for large-scale problems \cite{wang:djuric,djuric13a}. It is an SVM-like algorithm that formulates a non-linear model by assigning a number of linear hyperplanes to each class in order to capture data non-linearity. Given a $d$-dimensional example ${\bf x}$ and a set  $\mathcal{Y}$ of $L$ possible classes, AMM has the following form,
\begin{equation}
\label{eq1}
f({\bf x})=\arg \max_{i\in\mathcal{Y}}g(i, {\bf x}),
\end{equation}
where the scoring function $g(i, {\bf x})$ for the $i^{\text{\scriptsize th}}$ class,
\begin{equation}
\label{eq2}
g(i, {\bf x}) = \max_j {\bf w}_{i,j}^{\mathrm{T}} {\bf x},
\end{equation}
is parameterized by a weight matrix ${\bf W}$ written as
\begin{equation}\label{eq3}
{\bf W} = \bigg[{\bf w}_{1,1}\ldots{\bf w}_{1,b_1} | {\bf w}_{2,1}\ldots{\bf w}_{2,b_2} | \ldots |{\bf w}_{L,1} \ldots {\bf w}_{L,b_L}\bigg],
\end{equation}
where $b_1,\ldots,b_{L}$ are the numbers of weights (i.e., hyperplanes) assigned to each of the $L$ classes, and each block in (\ref{eq3}) is a set of class-specific weights. Thus, from (\ref{eq1}) we can see that the predicted label of the example ${\bf x}$ is the class of the weight vector that achieves the maximum value $g(i, {\bf x})$.

AMM is trained by minimizing the following convex problem at each $t^{\text{\scriptsize th}}$ training iteration,
\begin{equation}
\label{eq:loss_inst}
\mathcal{L}^{(t)} ({\bf{W}} |{\bf{z}}) \equiv \frac{\lambda }{2}||{\bf{W}}||_F^2 + l ({\bf{W}};({\bf{x}}_t ,y_t );z_t ),
\end{equation}
where $\lambda$ is the regularization parameter, and the instantaneous loss $l(\cdot)$ is computed as
\begin{equation}
\label{eq3.7.1}
 l ({\bf{W}};({\bf{x}}_t ,y_t );z_t ) = \max \left( {0,1 + \max _{i\in \mathcal{Y} \backslash y_t}g(i,{\bf{x}}_t ) - {\bf{w}}_{y_t ,z_t }^{\mathrm{T}} {\bf{x}}_t } \right).
\end{equation}
Element $z_t$ of vector ${\bf{z}} = [z_1^{} \ldots z_T^{} ]$ determines which weight belonging to the true class of the $t^{\text{\scriptsize th}}$ example is used to calculate (\ref{eq3.7.1}), and can be fixed prior to the start of a training epoch or, as done in this paper, can be computed on-the-fly as an index of a true-class weight that provides the highest score \cite{wang:djuric}.

AMM uses Stochastic Gradient Descent (SGD) to solve (\ref{eq:loss_inst}). The SGD is initialized with the zero-matrix (i.e., ${\bf W}^{(0)} = {\bf 0}$), which comprises infinite number of zero-vectors for each class.
 This is followed by an iterative procedure, where training examples are observed one by one and the weight matrix is modified accordingly. Upon receiving example $({\bf x}_t,y_t)\in \mathcal{D}$ at the $t^{\text{\scriptsize th}}$ round, ${\bf{w}}_{ij}^{(t)}$ is updated as
\begin{equation}\label{eq3.8}
{\bf{w}}_{ij}^{(t + 1)}  = {\bf{w}}_{ij}^{(t)}  - \eta ^{(t)} \nabla_{ij}^{(t)},
\end{equation}
where $\eta^{(t)}=1/(\lambda t)$ is a learning rate, and $\nabla_{ij}^{(t)}$ is the sub-gradient of (\ref{eq:loss_inst}) with respect to ${\bf{w}}_{i,j}^{(t)} $,
\begin{equation}
{\rm{ }}\nabla _{i,j}^{(t)}  = \left\{ \begin{array}{l}
 \lambda {\bf{w}}_{i,j}^{(t)}  + {\bf{x}}_t , \ \ {\text{   if~}}i = i_t^{} {\rm{,  }} \ j = j_t^{} {\rm{  }}, \\
 \lambda {\bf{w}}_{i,j}^{(t)}  - {\bf{x}}_t , \ \ {\text{   if~}}i = y_t^{} {\rm{, }} \ j = z_t^{} , \\
 \lambda {\bf{w}}_{i,j}^{(t)}, \ \ \ \ \ \ \ \ \ {\text{~~~otherwise,}} \\
 \end{array} \right.
\end{equation}
with
\begin{equation}
i_t^{}  = \mathop {\arg \max }\limits_{k \in {\cal Y}\backslash y_t
} g(k,{\bf{x}}) \ \ {\text{ and }} \ \ j_t^{}  = \mathop {\arg \max
}\limits_k ({\bf{w}}^{(t)}_{i_t ,k})^{\mathrm{T}} {\bf{x}}_t .
\end{equation}
If the loss (\ref{eq3.7.1}) at the $t^{\text{\scriptsize th}}$ iteration is positive, class weight from the true class $y_t$ indexed by $z_t$ is moved towards ${\bf x}_t$ during the update, while the class weight ${\bf{w}}_{i_t ,j_t }^{(t)}$ with the maximum prediction from the remaining classes is pushed away. If the updated weight is a zero-weight then it becomes non-zero, thus increasing the weight count $b_i$ for that class by one. In this way, complexity of the model adapts to complexity of the data, and $b_i, i = 1, \ldots, L$, are learned during training.

\section{Methodology}
It has been shown that the existing label ranking methods achieve good performance on many tasks, however, in the large-scale setting considered in this paper, they might not be as effective. When faced with non-linear problems comprising millions of examples and features, the proposed methods are either too costly to train and use, or may not be expressive enough to learn complex problems. To address this issue, in this section we present a novel ranking algorithm, called AMM-rank, that extends the idea of adaptability and online learning from AMM to label ranking setting, allowing large-scale training of accurate ranking models.

\subsection{AMM-rank algorithm}
Before detailing the training procedure of AMM-rank, we first consider its predictive label ranking model. As discussed previously, we assume that the $t^{\text{\scriptsize th}}$ training example ${\bf x}_t$ is associated with (possibly) incomplete label ranking $\pi_t$ of length $L_t \le L$. Given a trained AMM-rank model (\ref{eq3}) and a test example ${\bf x}$, a score for each class is found using equation (\ref{eq2}), and the predicted label ranking is obtained by sorting the scores in the descending order,
\begin{equation}
{\hat{\pi}} = {\rm{sort}}([g(1, {\bf x}), g(2, {\bf x}), \ldots, g(L, {\bf x})]),
\end{equation}
where the sort function returns indices of the sorted scores.

Training of AMM-rank resembles the training of AMM multi-class model described in the previous section. Learning is initialized with a zero-matrix comprising an infinite number of zero-vectors for each class, followed by iteratively observing examples one by one and modifying the weight matrix. At each $t^{\text{\scriptsize th}}$ training iteration we minimize the following regularized instanteneous rank loss,
\begin{equation}
\label{eq:loss_rank}
\mathcal{L}_{rank}^{(t)} ({\bf{W}} |{\bf{z}}) \equiv \frac{\lambda }{2}\|{\bf{W}}\|_F^2 + l_{rank} ({\bf{W}};({\bf{x}}_t ,y_t );{\bf z}_t ).
\end{equation}
The ranking loss $l_{rank}(\cdot)$ is defined as
\begin{equation}
\label{eq:loss_rank1}
\begin{aligned}
& l_{rank} ({\bf{W}};({\bf{x}}_t ,y_t );{\bf z}_t ) = \\
& \sum_{i = 1}^{L_t} \nu(i)  \sum_{j = 1}^{L}I(\pi_i \succ j) \max(0, 1 + g(j, {\bf{x}}_t ) - {\bf{w}}_{\pi_i, z_{ti} }^{\mathrm{T}} {\bf{x}}_t),
\end{aligned}
\end{equation}
where $\nu(i)$ is a predefined importance assigned to the $i^{\text{\scriptsize th}}$ rank, and function $I(arg)$ returns $1$ if $arg$ evaluates to true, and $0$ otherwise. As in label ranking setting we need to keep track of predicted scores of all $L$ classes and not only the top one, note that we introduced vector ${\bf z}_t$ instead of a scalar $z_t$ as in (\ref{eq:loss_inst}), whose element $z_{ti}$ determines which weight belonging to label $i$ is used to compute (\ref{eq:loss_rank}) for the $t^{\text{\scriptsize th}}$ example. 

Depending on the problem at hand, using the function $\nu(i)$ a modeler can emphasize the importance of some ranks over the others. For example, let us assume $\nu(i) = 1 / i$. Then, in the ranking loss defined in (\ref{eq:loss_rank1}), the factor $i^{-1}$ enforces higher penalty for misranking of top-ranked topics, while the mistakes made for lower-ranked topics incur progressively smaller costs. This approach has been explored previously in information retrieval setting \cite{weston2012}. However, it is also applicable in the context of targeted advertising, where lower-ranked classes have progressively lower relevance to an ad publisher than the higher-ranked ones. Furthermore, penalty is incurred whenever the lower-ranked label was either predicted to be preferred over the higher-ranked one, or the score of the preferred label was higher with a margin smaller than $1$. 

We use SGD at each training iteration to minimize the objective function (\ref{eq:loss_rank}). Subgradient of the instantaneous rank loss with respect to the weights can be computed as
\begin{equation}
\label{eq:sgd_up}
\begin{aligned}
& \nabla _{i,j}^{(t)}  = ~  \lambda {\bf{w}}_{i,j}^{(t)}  -  {\bf x}_t  ~ I(j = z_{ti}) ~ \nu(\pi_i^{-1}) \sum_{k = 1}^L \Big( I \big(i \succ k) \cdot \\
& ~~~~~I(1 + g(k, {\bf x}_t) > ({\bf w}_{ij}^{(t)})^{\mathrm{T}} {\bf x}_t \big) \Big) + {\bf x}_t~I(j = z_{ti})  \cdot \\
& ~~~~\sum_{k = 1}^L \Big( \nu(k)~ I(k \succ i) ~I \big(1 + ({\bf w}_{ij}^{(t)})^{\mathrm{T}} {\bf x}_t > ({\bf w}_{kz_{tk}}^{(t)})^{\mathrm{T}} {\bf x}_t \big) \Big).
\end{aligned}
\end{equation}
An SGD update step (\ref{eq:sgd_up}) can be summarized as follows. At every training round all model weights are reduced towards zero by multiplying them with $(1-1/t)$ (the first term on the RHS). In addition, if the $j^{\text{\scriptsize th}}$ weight of the $i^{\text{\scriptsize th}}$ class was used to compute the score for the $t^{\text{\scriptsize th}}$ label (i.e., $I(j = z_{ti})$ equals $1$), it is pushed further towards ${\bf x}_t$ whenever the $i^{\text{\scriptsize th}}$ label was either wrongly predicted to be less preferred or correctly predicted with margin smaller than $1$ (the second term on the RHS). Moreover, the weight is pushed further away from ${\bf x}_t$ whenever the score of the class preferred over the $i^{\text{\scriptsize th}}$ class was either lower or higher with margin less than $1$ (the third term on the RHS). Similarly to the AMM model, the complexity of AMM-rank ranking model is automatically learned during training, and adapts to the complexity of the considered label ranking problem.


\section{Experiments}
In this section we describe the problem setting and present a large-scale, real-world data set that was used for evaluation, followed by description and analysis of empirical results.

\subsection{Dataset}
We are addressing a problem from display advertising domain which consists of several key players: 1) advertisers, companies that want to advertise their products; 2) publishers, websites that host the advertisements (such as Yahoo or Google); and 3) online users. The web environment provides publishers with the means to track user behavior in much greater detail than in the offline setting, including capturing user's registered information (e.g., demographics, location) and activity logs that comprise search queries, page views, email activity, ad clicks, and purchases. This brings the ability to target users based on their past behavior, which is typically referred to as ad targeting \cite{ahmed2011scalable,pandey2011learning,tyler2011retrieval,agarwal2012targeting,aly2012web}. Having this in mind, the main motivation for the following experimental setup was the task of estimating user's ad click interests using their past activities. The idea is that, if we sort the interests in descending order of preference and attempt to predict this ranking, this task can be formulated as a label ranking problem.

The data set that was used in the empirical evaluation was generated using the information about users' online activities collected at Yahoo servers. The activities are temporal sequences of raw events that were extracted from server logs and are represented as tuples $(u_i, e_i, t_i), i = 1, \ldots, N$, where $u_i$ is ID of a user that generated the $i^{\text{\scriptsize th}}$ tuple, $e_i$ is an event type, $t_i$ is a timestamp, and $N$ is a total number of recorded tuples. For each user we considered events belonging to one of the following six groups:
\begin{itemize}
\item page views ("pv") - website pages that the user visited;
\item search queries ("sq") - user-generated search queries;
\item search link clicks ("slc") - user clicks on search links;
\item sponsored link clicks ("olc") - user clicks on search-advertising links that appear next to actual search links;
\item ad views ("adv") - display ads that the user viewed;
\item ad clicks ("adc") - display ads that the user clicked on.
\end{itemize}
Events from these six groups are all categorized into an in-house hierarchical taxonomy by an automatic categorization system and human editors. Each event is assigned to a category from a leaf of the taxonomy, and then propagated upwards toward parent categories. Considering that the server logs for each user are retained for several months, the recorded events can be used to capture users' interests in categories over long periods of time.


Following the ad categorization step, we can compute intensity and recency measures for each of $L$ considered categories in each of the six groups.  Let $\mathcal{D}_{ugct}$ denote a set of all tuples that were generated by user $u$, where $e_i$ belongs to group $g$ and is labeled with category $c$, with timestamp $t_i \le t$. Then, intensity and recency are defined as follows,
\begin{itemize}
\item {\bf intensity} is an exponentially time-decayed count of all tuples in $\mathcal{D}_{ugct}$, computed as
\begin{equation}
\label{eq:intense}
intensity(u, g, c, t) = \sum_{(u_i, e_i, t_i) \in \mathcal{D}_{ugct}} {\alpha}^{t-t_i},
\end{equation}
where $\alpha$ is a fixed decay factor, with $0 < \alpha < 1$ (we omit the exact value as it represents sensitive information).
\item {\bf recency} is a difference between timestamp $t$ and a timestamp of the most recent event from $\mathcal{D}_{ugct}$, computed as
\begin{equation}
\label{eq:recent}
recency(u, g, c, t) = \min_{(u_i, e_i, t_i) \in \mathcal{D}_{ugct}}(t - t_i).
\end{equation}
\end{itemize}

The intensity and recency measures were used to generate both the features and the label ranks for each user. In particular, we first chose two timestamps that were one month apart, $T_{features}$ and $T_{labels}$, where $T_{features} < T_{labels}$. Then, at timestamp $T_{features}$ we used (\ref{eq:intense}) and (\ref{eq:recent}) to compute intensity and recency of $L$ categories in each of "pv", "sq", "slc", and "olc" groups separately, which, together with user's age (split into $9$ buckets and represented as $9$ binary features) and gender (represented as $2$ binary features) was used as a feature vector ${\bf x}$, resulting in input space dimensionality $d$ of $(2\cdot4\cdot L + 9 + 2)$. In addition, in order to evaluate the influence of user ad views to their ad clicks, we also considered the case where intensity and recency of $L$ categories in the "adv" group were appended to the feature vector, increasing the dimensionality by $2L$.

Furthermore, to quantify user interests and generate ground-truth ranks $\pi$, we considered only "adc" events between $T_{features}$ and $T_{labels}$, and computed intensity of $L$ categories at timestamp $T_{labels}$. We consider level of interest of user $u$ in category $c$ to be equal to intensity of $c$ in "adc" group, and preference ranking of categories is obtained simply by sorting their intensities. Note that the ground-truth ranking is in most cases incomplete, as users usually do not interact with all categories from the taxonomy.

\begin{figure}[t]
\centering 
\includegraphics[width=0.40\textwidth]{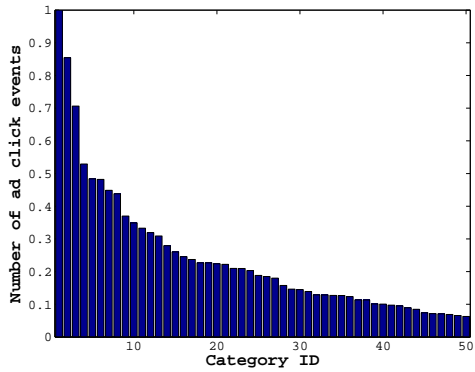}
\caption{Number of ad click events per category}
\label{fig:hist}
\end{figure}

We considered $L = 50$ second-level categories of the taxonomy (e.g., "finance/loans", "retail/apparel"), and collected data comprising $3{,}289{,}229$ anonymous users that clicked on more than $2$ categories. Category distribution in the ground-truth ranks is given in Fig. \ref{fig:hist}, where we see that a large fraction of ad clicks would be missed if users were targeted only with the most clicked categories, which directly results in lost revenue for both publishers and advertisers.

\begin{figure}[!t]
    \centering
    \adjustbox{minipage=7cm,margin=1em,width=0.36\textwidth,frame,center}{
{\bf Females, aged 21-25}\\
01.	Retail/Apparel\\
02.	Technology/Internet Services\\
03.	Telecommunications/Cellular and Wireless\\
04.	Travel/Destinations\\
05.	Consumer Goods/Beauty and Personal Care\\
06.	Technology/Consumer Electronics\\
07.	Consumer Goods/Contests and Sweepstakes\\
08.	Travel/Vacations\\
09.	Travel/Non US\\
10.	Life Stages/Education\\
\\
{\bf Males, aged 21-25}\\
01.	Technology/Internet Services\\
02.	Retail/Apparel\\
03.	Telecommunications/Cellular and Wireless\\
04.	Travel/Destinations\\
05.	Technology/Consumer Electronics\\
06.	Travel/Non US\\
07.	Travel/Vacations\\
08.	Consumer Goods/Contests and Sweepstakes\\
09.	Retail/Home\\
10.	Entertainment/Games\\
\\
{\bf Females, aged 65+}\\
01.	Consumer Goods/Beauty and Personal Care\\
02.	Retail/Apparel\\
03.	Life Stages/Education\\
04.	Finance/Loans\\
05.	Finance/Insurance\\
06.	Finance/Investment\\
07.	Technology/Internet Services\\
08.	Entertainment/Television\\
09.	Retail/Home\\
10.	Telecommunications/Cellular and Wireless\\
\\
{\bf Males, aged 65+}\\
01.	Finance/Investment\\
02.	Finance/Loans\\
03.	Retail/Apparel\\
04.	Life Stages/Education\\
05.	Technology/Internet Services\\
06.	Finance/Insurance\\
07.	Consumer Goods/Beauty and Personal Care\\
08.	Retail/Home\\
09.	Telecommunications/Cellular and Wireless\\
10.	Technology/Computer Software\\
}
    \caption{Topic ranking found by the AG-Mal model}
    \label{fig:borda}
\end{figure}

\subsection{Results}
We compared AMM-rank to following approaches: a) multi-class AMM \cite{wang:djuric}, where the top-ranked category was used as a true class and the output scores for all categories were sorted to obtain ranking, used as a na\"{\i}ve baseline; b) Central-Mal, which always predicts central ranking of the training set computed using the Mallows model \cite{mallows1957non}; c) AG-Mal, which computes Central-Mal over all users grouped in different age ("13-17", "18-20", "21-24", "25-29", "30-34", "35-44", "45-54", "55-64", "65+") and gender (male/female) buckets; d) IB-Mal, which computes Central-Mal over $k$ nearest neighbors \cite{Cheng1}; e) logistic regression (LR), where $L$ binary models were trained and we sorted their outputs to obtain a ranking; and f) pairwise approach \cite{hullermeier2008label}, where  $L (L-1)/2$ binary LR models were trained and we sorted the sum of their soft votes towards each label to obtain a ranking (PW-LR). AMM-rank and PW-LR have $\mathcal{O}(NL^2)$ and IB-Mal has $\mathcal{O}(N^2L)$ time complexity, while the remaining methods are $\mathcal{O}(NL)$ approaches.

Central-Mal is a very simple and efficient baseline, and is an often-used method for basic content personalization. As the method simply predicts population's mean ranking, to improve its performance we considered AG-Mal, a method commonly used in practice, where we first compute mean rank for each age-gender group, and then use the group's mean rank as a prediction for qualified users. Further, IB-Mal is an instance-based method which is extremely competitive to the other state-of-the-art approaches (e.g., see Grbovic, Djuric, and Vucetic 2013), where we first find $k$ nearest neighbors by considering feature vectors ${\bf x}$ and then predict Mallows mean ranking over the neighbors (due to large time cost, for each user we search for nearest neighbors in a subsampled set of $100{,}000$ users). Lastly, we considered LR since it represents industry standard for ad targeting tasks, and PW-LR as it was shown to achieve state-of-the-art performance on a number of ranking tasks \cite{grbovic2012supervised,grbovic2013}. Due to large scale of the problem, we did not consider state-of-the-art methods such as mixture models which require iterative training \cite{grbovic2012learning,grbovic2013supervised}. We also did not consider log-linear model \cite{Dekel}, shown in \cite{grbovic2013} to be outperformed by the IB-Mal, and do not report results of instance-based Plackett-Luce \cite{Cheng2} due to observed limited performance.

We used Vowpal Wabbit package\footnote{\url{github.com/JohnLangford/vowpal_wabbit}} for logistic regression, BudgetedSVM \cite{djuric13a} for AMM, that we also modified to implement AMM-rank. We set $\nu(i) = 1, i = 1, \ldots, L$, and used the default parameters from BudgetedSVM package for AMM-rank, with the exception of the $\lambda$ parameter which, together with competitors' parameters, was configured through cross-validation on a small held-out set; this resulted in $k = 10$ for IB-Mal.  
As discussed previously, we considered two versions of the ad targeting data:
\begin{itemize}
\item {\bf \st{adv}} - feature vector ${\bf x}$ does not include recency and intensity of categories from "adv" group (with $d = 411$);
\item {\bf adv}, feature vector ${\bf x}$ does include recency and intensity of categories from "adv" group (with $d = 511$).
\end{itemize}

\begin{figure*}[t]
\centering 
\includegraphics[width=0.33\textwidth]{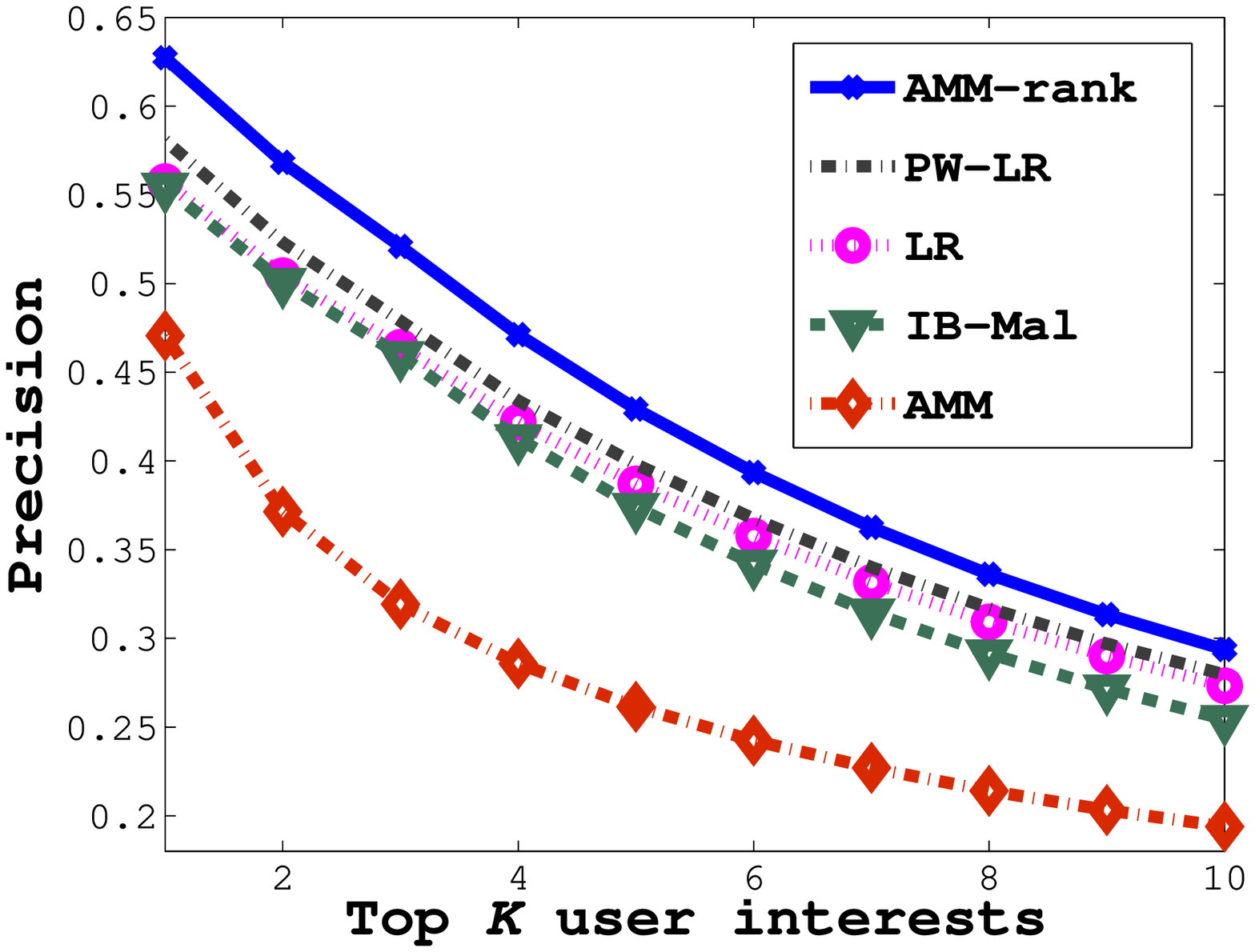}
\includegraphics[width=0.33\textwidth]{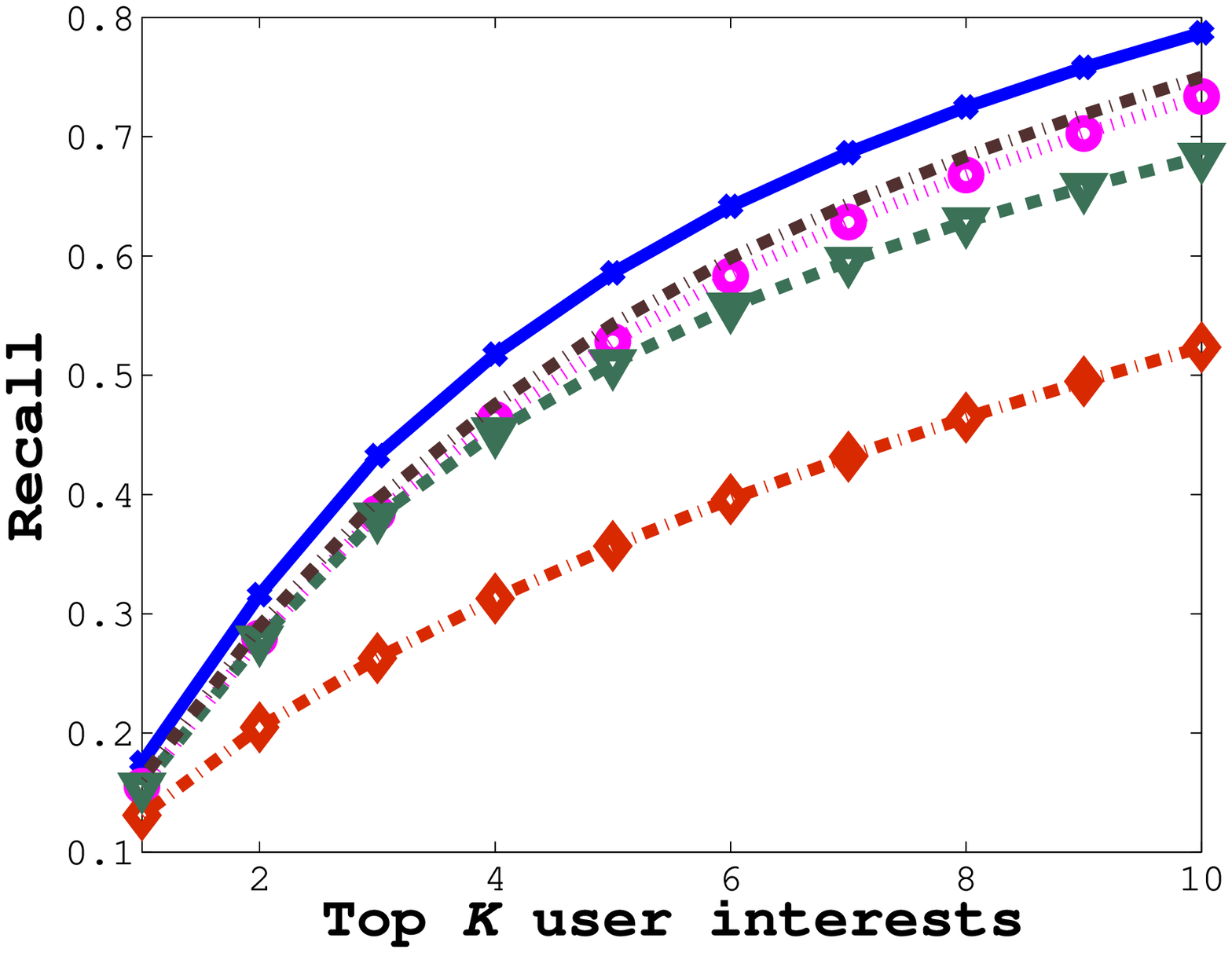}
\includegraphics[width=0.33\textwidth]{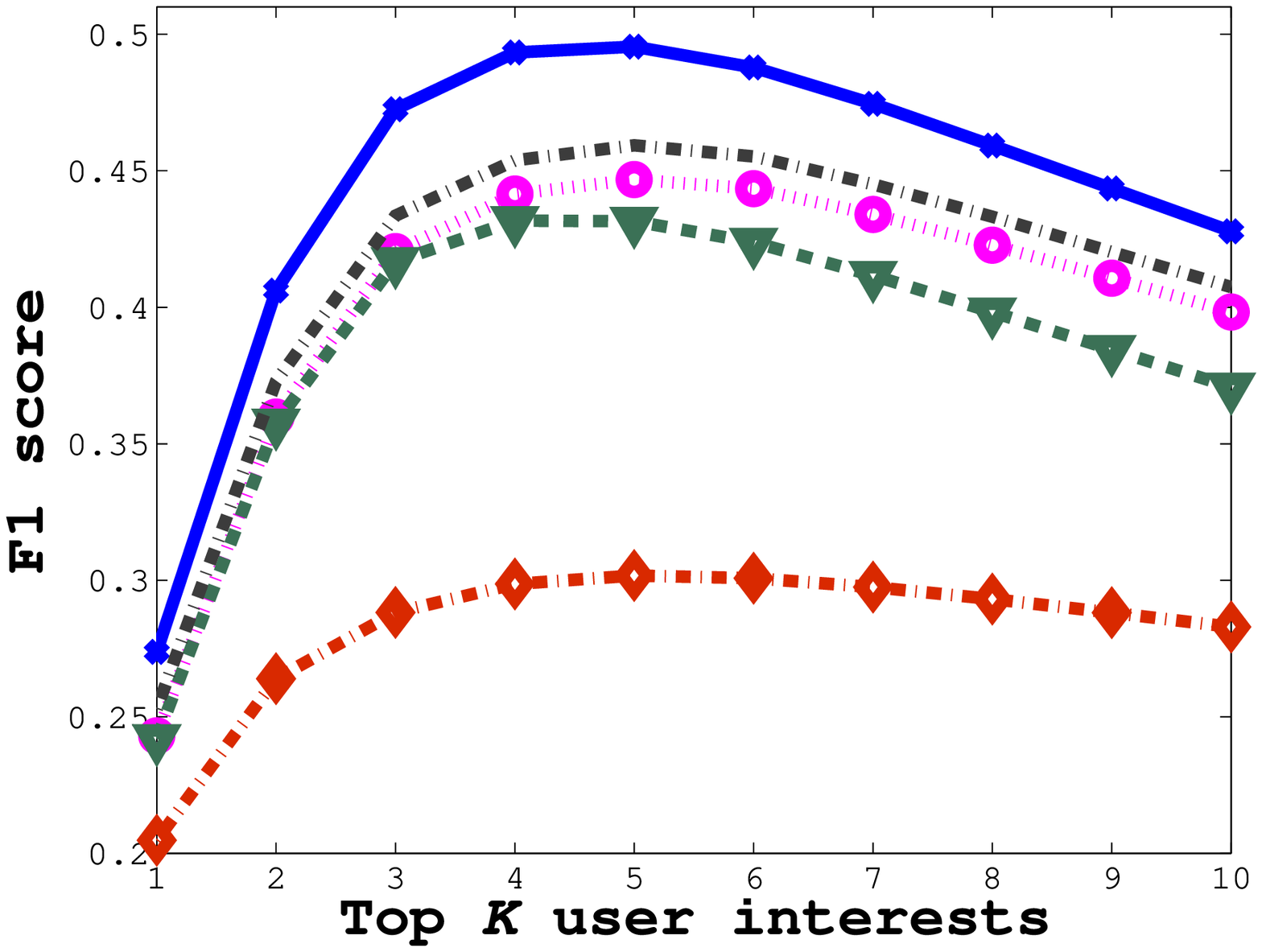}
\caption{Comparison of retrieval performance of label ranking algorithms in terms of precision, recall, and F1 measures}
\label{fig:prec}
\end{figure*}

\begin{table}
\caption{Disagreement error $\epsilon_{\text{dis}}$ of the label ranking methods}
\label{table_disagreement_loss}
\begin{center}
\begin{tabular}{lcccc}
\rule{0pt}{2.5ex}{\bf Algorithm} & {\bf \st{adv}} & {\bf adv}  \\
\hline \hline
\rowcolor{lightgray}
AMM  & 0.3446& 0.2611\\
Central-Mal & 0.2957  & 0.2957\\ 
\rowcolor{lightgray}
AG-Mal  & 0.2820 & 0.2820\\
IB-Mal  & 0.2694 & 0.1899\\
\rowcolor{lightgray}
LR & 0.2110 & 0.1419\\
PW-LR & 0.2091 & 0.1226\\
\rowcolor{lightgray}
AMM-rank & \textbf{0.1996}  & \textbf{0.1083} \\ 
 \bottomrule
\end{tabular}
\end{center}
\end{table}

Before comparing the ranking approaches, it is informative to consider the examples of label ranks found by AG-Mal on {\bf adv} data, given in Figure \ref{fig:borda}. We can see that there exist significant differences between different gender and age groups. Albeit the obtained ranks seem very intuitive, we will see shortly that AG-Mal is significantly outperformed by the other methods, illustrating complexity of the ranking task and the need for more involved approaches. In the following, we compare the algorithms using disagreement error $\epsilon_{\text{dis}}$ \cite{Dekel}, computed as a fraction of pairwise category preferences predicted incorrectly,
\begin{equation}
\label{eq:dis_loss}
\begin{aligned}
\epsilon_{\text{dis}} = \frac{1}{N_{test}}\sum_{t=1}^{N_{test}}\sum_{i,j=1}^{L}\frac{I\big((\pi_{ti} \succ \pi_{tj}) \wedge (\hat{\pi}_{t\pi_{tj}}^{-1} > \hat{\pi}_{t\pi_{ti}}^{-1})\big)}{L_t \big(L - 0.5 (L_t + 1)\big)},
\end{aligned}
\end{equation}
as well as precision, recall, and F1 at the top $K$ ranks,
\begin{equation}
\label{eq:measures}
\begin{aligned}
\text{precision@}K &= \frac{1}{N_{test}}\sum_{t=1}^{N_{test}}\sum_{i=1}^{K}\frac{I(\hat{\pi}_{ti} \in \pi_t)}{K},\\
\text{recall@}K &= \frac{1}{N_{test}}\sum_{t=1}^{N_{test}}\sum_{i=1}^{K}\frac{I(\hat{\pi}_{ti} \in \pi_t)}{L_t},\\
\text{F$1$@}K &= \frac{2 \cdot \text{precision@}K \cdot \text{recall@}K}{\text{precision@}K + \text{recall@}K},
\end{aligned}
\end{equation}
which are commonly used measures for ranking problems. Here, $\hat{\pi}_t$ denotes predicted label rank for the $t^{\text{\scriptsize th}}$ example.

Performance of the competing methods in terms of $\epsilon_{\text{dis}}$, following $5$-fold cross-validation, is reported in Table \ref{table_disagreement_loss}. We can see that the inclusion of ad view features resulted in large performance improvement, confirming findings from \cite{gupta2012factoring} that past exposure to an ad increases propensity of a user to actually click the ad.
As expected, multi-class AMM achieved poor performance as it optimizes only for the topmost category, and this result represents a lower bound on the disagreement loss. A simple baseline Central-Mal achieved higher error, which was decreased by only a small margin using AG-Mal. We can see that IB-Mal resulted in significant performance improvement, however in large-scale, online setting it may be very inefficient. Logistic regression, a commonly used method in ad targeting tasks, obtained low error, further improved using the pairwise approach. However, state-of-the-art PW-LR was significantly outperformed by the proposed AMM-rank which achieved more than $10\%$ better result. We note that, other than IB-Mal, the methods are very efficient, obtaining training and test times of less than $10$ minutes on a regular machine.

However, the main goal in ad targeting campaigns is not to infer the complete list of preferences for a user. Instead, we aim to find the top $K$ most preferred categories, due to the constraint that we only have a limited budget for ad display, in terms of both time and space. Therefore, it is not of importance when two less preferred categories are misranked, and in the second set of experiments we explore how the label ranking methods perform in such setting. We considered showing $K = \{1, 2, \ldots, 10\}$ display ads, and for the top $K$ ranks measure precision, recall, and F$1$ score of the categories on which the user clicked during the testing period. The results obtained by the label ranking algorithms are illustrated in Figure \ref{fig:prec}. We can see that AMM-rank outperformed the competitors, achieving better performance for all values of $K$. This becomes even more relevant when we consider that even a small improvement in a web-scale setting of targeted advertising may result in a significant revenue increase for the publisher. We can conclude that the results strongly suggest advantages of the proposed approach over the competing algorithms in large-scale label ranking tasks. 

\section{Conclusion}
In order to address challenges brought about by the scale of the online advertising tasks that renders many state-of-the-art methods inefficient, we introduced AMM-rank, a novel, non-linear algorithm for large-scale label ranking. We evaluated its performance on a real-world ad targeting data comprising more than 3 million users, thus far the largest label ranking data considered in the literature. The results show that the method outperformed the competing approaches by a large margin in terms of both rank loss and retrieval measures, indicating that the AMM-rank algorithm is a very suitable method for solving large-scale label ranking problems.

\balance
%
\bibliographystyle{aaai}
\bibliography{refs}  

\begin{thebibliography}{}

\bibitem[\protect\citeauthoryear{Agarwal, Pandey, and
  Josifovski}{2012}]{agarwal2012targeting}
Agarwal, D.; Pandey, S.; and Josifovski, V.
\newblock 2012.
\newblock Targeting converters for new campaigns through factor models.
\newblock In {\em Proceedings of the 21st International Conference on World
  Wide Web},  101--110.
\newblock ACM.

\bibitem[\protect\citeauthoryear{Ahmed \bgroup et al\mbox.\egroup
  }{2011}]{ahmed2011scalable}
Ahmed, A.; Low, Y.; Aly, M.; Josifovski, V.; and Smola, A.~J.
\newblock 2011.
\newblock Scalable distributed inference of dynamic user interests for
  behavioral targeting.
\newblock In {\em KDD},  114--122.

\bibitem[\protect\citeauthoryear{Alba \bgroup et al\mbox.\egroup
  }{1997}]{alba1997interactive}
Alba, J.; Lynch, J.; Weitz, B.; Janiszewski, C.; Lutz, R.; Sawyer, A.; and
  Wood, S.
\newblock 1997.
\newblock Interactive home shopping: consumer, retailer, and manufacturer
  incentives to participate in electronic marketplaces.
\newblock {\em The Journal of Marketing}  38--53.

\bibitem[\protect\citeauthoryear{Aly \bgroup et al\mbox.\egroup
  }{2012}]{aly2012web}
Aly, M.; Hatch, A.; Josifovski, V.; and Narayanan, V.~K.
\newblock 2012.
\newblock Web-scale user modeling for targeting.
\newblock In {\em WWW},  3--12.
\newblock ACM.

\bibitem[\protect\citeauthoryear{Broder}{2008}]{broder2008computational}
Broder, A.~Z.
\newblock 2008.
\newblock Computational advertising and recommender systems.
\newblock In {\em Proceedings of the ACM conference on Recommender systems},
  1--2.
\newblock ACM.

\bibitem[\protect\citeauthoryear{Cao \bgroup et al\mbox.\egroup
  }{2007}]{LearningToRank}
Cao, Z.; Qin, T.; Liu, T.; Tsai, M.; and Li, H.
\newblock 2007.
\newblock Learning to rank: From pairwise approach to listwise approach.
\newblock {\em ICML}  129--136.

\bibitem[\protect\citeauthoryear{Chellappa and
  Sin}{2005}]{chellappa2005personalization}
Chellappa, R.~K., and Sin, R.~G.
\newblock 2005.
\newblock {Personalization versus privacy: An empirical examination of the
  online consumerÕs dilemma}.
\newblock {\em Information Technology and Management} 6(2-3):181--202.

\bibitem[\protect\citeauthoryear{Cheng, Dembczy{\'n}ski, and
  H\"ullermeier}{2010}]{Cheng2}
Cheng, W.; Dembczy{\'n}ski, K.; and H\"ullermeier, E.
\newblock 2010.
\newblock {Label ranking methods based on the Plackett-Luce model}.
\newblock In {\em Proceedings of the 27th International Conference on Machine
  Learning},  215--222.

\bibitem[\protect\citeauthoryear{Cheng, H\"uhn, and
  H\"ullermeier}{2009}]{Cheng1}
Cheng, W.; H\"uhn, J.; and H\"ullermeier, E.
\newblock 2009.
\newblock {Decision tree and instance-based learning for label ranking}.
\newblock In {\em Proceedings of the 26th International Conference on Machine
  Learning},  161--168.

\bibitem[\protect\citeauthoryear{Das \bgroup et al\mbox.\egroup
  }{2007}]{das2007google}
Das, A.~S.; Datar, M.; Garg, A.; and Rajaram, S.
\newblock 2007.
\newblock {Google news personalization: Scalable online collaborative
  filtering}.
\newblock In {\em WWW},  271--280.
\newblock ACM.

\bibitem[\protect\citeauthoryear{Dekel, Manning, and Singer}{2003}]{Dekel}
Dekel, O.; Manning, C.; and Singer, Y.
\newblock 2003.
\newblock {Log-linear models for label ranking}.
\newblock In {\em Advances in Neural Information Processing Systems},
  volume~16. MIT Press.

\bibitem[\protect\citeauthoryear{Djuric \bgroup et al\mbox.\egroup
  }{2014}]{djuric13a}
Djuric, N.; Lan, L.; Vucetic, S.; and Wang, Z.
\newblock 2014.
\newblock {BudgetedSVM: A toolbox for scalable SVM approximations}.
\newblock {\em Journal of Machine Learning Research} 14:3813--3817.

\bibitem[\protect\citeauthoryear{Elisseeff and
  Weston}{2001}]{elisseeff2001kernel}
Elisseeff, A., and Weston, J.
\newblock 2001.
\newblock A kernel method for multi-labelled classification.
\newblock In {\em Advances in Neural Information Processing Systems},
  681--687.

\bibitem[\protect\citeauthoryear{Essex}{2009}]{essex2009matchmaker}
Essex, D.
\newblock 2009.
\newblock Matchmaker, matchmaker.
\newblock {\em Communications of the ACM} 52(5):16--17.

\bibitem[\protect\citeauthoryear{Grbovic \bgroup et al\mbox.\egroup
  }{2013}]{grbovic2013supervised}
Grbovic, M.; Djuric, N.; Guo, S.; and Vucetic, S.
\newblock 2013.
\newblock Supervised clustering of label ranking data using label preference
  information.
\newblock {\em Machine Learning}  1--35.

\bibitem[\protect\citeauthoryear{Grbovic, Djuric, and
  Vucetic}{2012a}]{grbovic2012learning}
Grbovic, M.; Djuric, N.; and Vucetic, S.
\newblock 2012a.
\newblock {Learning from pairwise preference data using Gaussian mixture
  model}.
\newblock {\em Preference Learning: Problems and Applications in AI}  33Ð--35.

\bibitem[\protect\citeauthoryear{Grbovic, Djuric, and
  Vucetic}{2012b}]{grbovic2012supervised}
Grbovic, M.; Djuric, N.; and Vucetic, S.
\newblock 2012b.
\newblock Supervised clustering of label ranking data.
\newblock In {\em SDM},  94--105.
\newblock SIAM.

\bibitem[\protect\citeauthoryear{Grbovic, Djuric, and
  Vucetic}{2013}]{grbovic2013}
Grbovic, M.; Djuric, N.; and Vucetic, S.
\newblock 2013.
\newblock Multi-prototype label ranking with novel pairwise-to-total-rank
  aggregation.
\newblock In {\em Proceedings of the 23rd International Joint Conference on
  Artificial Intelligence}.
\newblock AAAI Press.

\bibitem[\protect\citeauthoryear{Gupta \bgroup et al\mbox.\egroup
  }{2012}]{gupta2012factoring}
Gupta, N.; Das, A.; Pandey, S.; and Narayanan, V.~K.
\newblock 2012.
\newblock Factoring past exposure in display advertising targeting.
\newblock In {\em Proceedings of the ACM SIGKDD International Conference on
  Knowledge Discovery and Data mining},  1204--1212.

\bibitem[\protect\citeauthoryear{Har-Peled, Roth, and Zimak}{2003}]{Har-Peled}
Har-Peled, S.; Roth, D.; and Zimak, D.
\newblock 2003.
\newblock {Constraint classification for multiclass classification and
  ranking}.
\newblock In {\em Proceedings of the 16th Annual Conference on Neural
  Information Processing Systems},  785--792.
\newblock MIT Press.

\bibitem[\protect\citeauthoryear{H{\"u}llermeier and
  Vanderlooy}{2010}]{hullermeier2010combining}
H{\"u}llermeier, E., and Vanderlooy, S.
\newblock 2010.
\newblock {Combining predictions in pairwise classification: An optimal
  adaptive voting strategy and its relation to weighted voting}.
\newblock {\em Pattern Recognition} 43(1):128--142.

\bibitem[\protect\citeauthoryear{H{\"u}llermeier \bgroup et al\mbox.\egroup
  }{2008}]{hullermeier2008label}
H{\"u}llermeier, E.; F{\"u}rnkranz, J.; Cheng, W.; and Brinker, K.
\newblock 2008.
\newblock Label ranking by learning pairwise preferences.
\newblock {\em Artificial Intelligence} 172(16):1897--1916.

\bibitem[\protect\citeauthoryear{Kamishima and Akaho}{2006}]{Kamishima}
Kamishima, T., and Akaho, S.
\newblock 2006.
\newblock {Efficient clustering for orders}.
\newblock In {\em ICDM Workshops},  274--278.

\bibitem[\protect\citeauthoryear{Majumder and
  Shrivastava}{2013}]{majumder2013know}
Majumder, A., and Shrivastava, N.
\newblock 2013.
\newblock Know your personalization: Learning topic level personalization in
  online services.
\newblock In {\em Proceedings of the 22nd International Conference on World
  Wide Web},  873--884.

\bibitem[\protect\citeauthoryear{Mallows}{1957}]{mallows1957non}
Mallows, C.~L.
\newblock 1957.
\newblock Non-null ranking models.
\newblock {\em Biometrika} 44(1/2):114--130.

\bibitem[\protect\citeauthoryear{Manber, Patel, and
  Robison}{2000}]{manber2000yahoo}
Manber, U.; Patel, A.; and Robison, J.
\newblock 2000.
\newblock {Experience with personalization on Yahoo!}
\newblock {\em Communications of the ACM} 43(8):35.

\bibitem[\protect\citeauthoryear{Pandey \bgroup et al\mbox.\egroup
  }{2011}]{pandey2011learning}
Pandey, S.; Aly, M.; Bagherjeiran, A.; Hatch, A.; Ciccolo, P.; Ratnaparkhi, A.;
  and Zinkevich, M.
\newblock 2011.
\newblock Learning to target: what works for behavioral targeting.
\newblock In {\em Proceedings of the 20th ACM International Conference on
  Information and Knowledge Management},  1805--1814.
\newblock ACM.

\bibitem[\protect\citeauthoryear{Riecken}{2000}]{riecken2000personalized}
Riecken, D.
\newblock 2000.
\newblock Personalized views of personalization.
\newblock {\em Communications of the ACM} 43(8):27--28.

\bibitem[\protect\citeauthoryear{Tuzhilin}{2009}]{tuzhilin2009personalization}
Tuzhilin, A.
\newblock 2009.
\newblock {Personalization: The state of the art and future directions}.
\newblock {\em Business Computing} 3:3.

\bibitem[\protect\citeauthoryear{Tyler \bgroup et al\mbox.\egroup
  }{2011}]{tyler2011retrieval}
Tyler, S.~K.; Pandey, S.; Gabrilovich, E.; and Josifovski, V.
\newblock 2011.
\newblock Retrieval models for audience selection in display advertising.
\newblock In {\em Proceedings of the 20th ACM International Conference on
  Information and Knowledge Management},  593--598.
\newblock ACM.

\bibitem[\protect\citeauthoryear{Vembu and G{\"a}rtner}{2011}]{vembu2011label}
Vembu, S., and G{\"a}rtner, T.
\newblock 2011.
\newblock {Label ranking algorithms: A survey}.
\newblock In {\em Preference learning}. Springer.
\newblock  45--64.

\bibitem[\protect\citeauthoryear{Wang \bgroup et al\mbox.\egroup
  }{2011}]{wang:djuric}
Wang, Z.; Djuric, N.; Crammer, K.; and Vucetic, S.
\newblock 2011.
\newblock Trading representability for scalability: Adaptive multi-hyperplane
  machine for nonlinear classification.
\newblock In {\em ACM SIGKDD Conf. on Knowledge Discovery and Data Mining}.

\bibitem[\protect\citeauthoryear{Weston \bgroup et al\mbox.\egroup
  }{2012}]{weston2012}
Weston, J.; Wang, C.; Weiss, R.; and Berenzweig, A.
\newblock 2012.
\newblock Latent collaborative retrieval.
\newblock In {\em Proceedings of the 29th International Conference on Machine
  Learning},  9--16.

\end{thebibliography}
\balance

\end{document}